\documentclass[letterpaper, 10 pt, conference]{ieeeconf-modified}  

\IEEEoverridecommandlockouts                              

\usepackage{amsmath} 
\usepackage{amssymb}  
\usepackage{siunitx}
\usepackage[caption=false,font=footnotesize]{subfig}
\usepackage{tikz}
\usepackage{tkz-euclide}
\usepackage[numbers]{natbib}
\usepackage{layouts}
\usepackage{mathtools}
\usepackage{hyperref}
\usepackage[capitalize]{cleveref}
\usepackage{pgf}
\usepackage{dsfont}
\newcommand{\comment}[1]{}
\newcommand{\degree}{$^{\circ}$}

\newcommand{\norm}[1]{\left\lVert#1\right\rVert}
\newcommand{\abs}[1]{\lvert#1\rvert}
\newcommand{\ltwo}[1]{\left\lVert#1\right\rVert^2}

\graphicspath{{graphics/}, {figures/}}
\usetikzlibrary{shadows.blur,calc,matrix,chains,positioning,decorations.pathreplacing,arrows,shapes.geometric}
\crefname{equation}{}{}  

\title{\LARGE \bf
A Learning-based Controller for Multi-Contact Grasps \\on Unknown Objects with a Dexterous Hand
}

\author{Dominik Winkelbauer$^{1,2}$, Rudolph Triebel$^{1, 3}$, Berthold Bäuml$^{1,2}$
\thanks{$^{1}$DLR Institute of Robotics \& Mechatronics, Germany}%
\thanks{$^{2}$Learning AI for Dextrous Robots Lab (aidx- lab.org), Technical University of Munich, Germany}%
\thanks{$^{3}$Karlsruhe Institute of Technology (KIT), Germany}%
\thanks{Contact: \tt\footnotesize{Dominik.Winkelbauer@dlr.de}}%
\thanks{*This work is supported by the Helmholtz Association under the joint research school “Munich School for Data Science - MUDS“.}
}

\begin{document}

\maketitle
\thispagestyle{empty}
\pagestyle{empty}


\begin{abstract}
Existing grasp controllers usually either only support finger-tip grasps or need explicit configuration of the inner forces.
We propose a novel grasp controller that supports arbitrary grasp types, including power grasps with multi-contacts, while operating self-contained on before unseen objects. 
No detailed contact information is needed, but only a rough 3D model, e.g., reconstructed from a single depth image.
First, the external wrench being applied to the object is estimated by using the measured torques at the joints.
Then, the torques necessary to counteract the estimated wrench while keeping the object at its initial pose are predicted. 
The torques are commanded via desired joint angles to an underlying joint-level impedance controller.
To reach real-time performance, we propose a learning-based approach that is based on a wrench estimator- and a torque predictor neural network.
Both networks are trained in a supervised fashion using data generated via the analytical formulation of the controller.
In an extensive simulation-based evaluation, we show that our controller is able to keep \SI{83.1}{\percent}
 of the tested grasps stable when applying external wrenches with up to \SI{10}{\newton}.
At the same time, we outperform the two tested baselines by being more efficient and inducing less involuntary object movement.
Finally, we show that the controller also works on the real DLR-Hand II, reaching a cycle time of \SI{6}{\milli\second}.
Website: \href{https://aidx-lab.org/grasping/iros24}{aidx-lab.org/grasping}
\end{abstract}
 
\section{Introduction}

Robust grasping is one of the most important skills a robot needs to be able to interact with the world.
Planning the grasp, meaning positioning the hand and the fingers such that a stable grasp originates, has been the focus of extensive research~\cite{ Lu2020, Wu2020, Aktas2019, Mandikal2021, Winkelbauer2022}.
However, especially when using multi-fingered hands, setting the correct torques to apply at each joint is also crucial for the stability of the grasp.
These torques need to be adapted to changing external wrenches and, at the same time, the torques should only be as large as necessary, to be energy-efficient and to keep the load on object and hand as small as possible.
Existing approaches that control the torques depending on a given grasping situation usually only support fingertip grasps where the fingers are in a non-singular position and the contact positions and normals are known~\cite{Schneider1992, Wimboeck2006,Wimbock2012, Pfanne2020}.
However, in reality, the opposite is usually the case:
Many grasps, especially power grasps, lead to multi-contacts across multiple links.
The object might be in contact with the fixed hand base or fingers might be in singular configurations.
Also, detailed contact information is not available.
In this work, we are approaching the grasp control problem without any constraints on the grasp or object.
We propose a two-step approach, shown in \cref{fig:intro}:
First, based on the measured torques, the external wrench acting on the object gets estimated.
Then, the desired joint angles given to the underlying joint-level impedance controller are adjusted to counter the estimated wrench.
Our controller can handle unknown objects without any human supervision and only uses the available sensory information, which includes the measured torques at each joint, the measured joint angles and the object observed via a depth camera.
By using learning-based methods, we are able to run the controller in real-time and do not need any explicit detailed contact information.
In summary, we make the following novel contributions:

\begin{figure}[t]
     \centering
     \includegraphics[width=0.47\columnwidth]{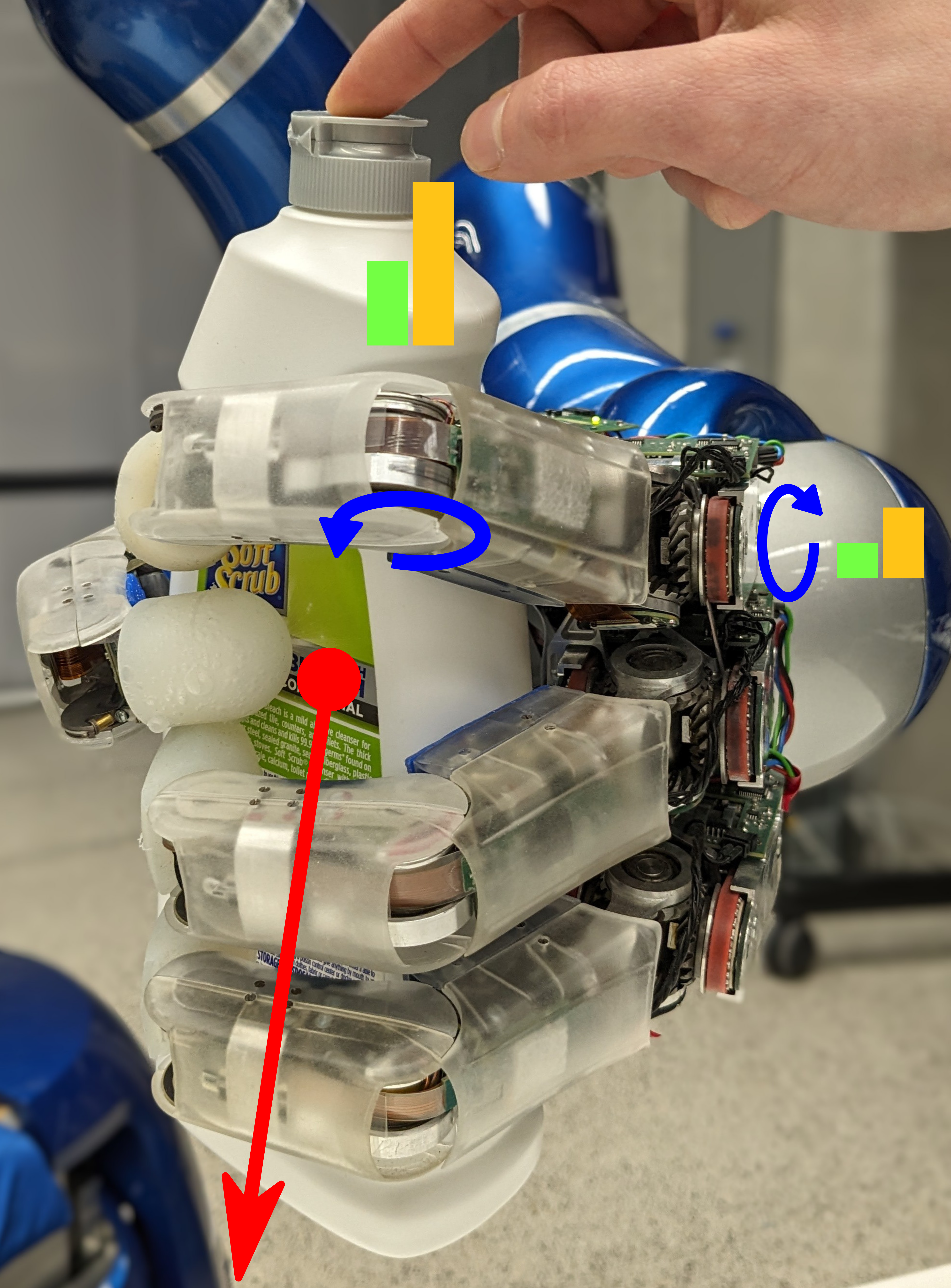}
         
     \caption{The controller applied to a grasp on the YCB bleach bottle. The applied external wrench (red arrow) gets estimated and then the commended torques are adjusted. Here, two joints are visualized with their applied torque before (green) and after (orange) the external wrench is applied.}
     \label{fig:intro}
\vspace{-15pt}
\end{figure}

\begin{itemize}
\item We propose a novel grasp controller that is able to estimate external wrenches using the measured joint torques and at the same time able to adapt the torques such that the wrench is countered
\item We provide an analytical formulation of the controller based on an elastic model that is able to handle all kinds of grasps including power grasps with multi-contacts and grasps in singular positions
\item We further propose a learning-based formulation that allows running the controller in real-time and without detailed contact information
\item To be able to handle ambiguous situations, we propose to train the wrench estimation network using a novel loss which makes the network predict always the smallest wrench explaining a given situation
\end{itemize}

\section{Related Work}

Grasping unknown objects using multi-fingered hands is becoming more and more popular, especially using data-driven methods.
While also the correct configuration of the grasp forces is a crucial part of the grasping process~\cite{Aktas2019}, most methods rely on a simple heuristic, e.g. apply the same amount of force in each finger~\cite{ Lu2020, Wu2020, Aktas2019, Mandikal2021}.
We clearly outperform such heuristics by being more efficient and inducing less undesired object movement.

There exist methods that provide a more sophisticated way of controlling grasps.
One of the most prominent approaches is the \textit{object-level impedance controller}, introduced first by \citet{Schneider1992}.
Usually, such controllers are combined with the virtual linkage model~\cite{Williams1993}, to be able to define the internal forces without having explicit knowledge of the contacts.
The concept of the object-level impedance controller has been later extended with a novel damping design and to not require a simulation of a virtual object~\cite{Wimbock2012}. 
However, the methods mentioned above require the user to configure the virtual linkage based on the grasped object.
Instead, our proposed method sets internal forces autonomously for unknown objects.

\citet{Pfanne2020} proposed to use vision-based contact detection which allows determining the correct internal forces.
However, their method only supports fingertip grasps, which is a general issue with object-level impedance controllers:
They require the object to be able to move in each direction, however, in power grasps not much object movement is possible.
In our approach, we only rely on the measured joint torques, which requires only little object movement.

Another way of object-stability control in grasping is by using a tactile sensor.
Here, the main idea is usually to increase the inner forces as soon as a slip is detected~\cite{Su2015, Wettels2009, Romano2011, Deng2020}.
Again, these methods only support fingertip grasps, because the tactile sensor is usually only available at the fingertip and the proposed adaptation of the inner forces does not support multi-contacts.

\citet{Bicchi1991} recognized that in power grasps the methods used for fingertip grasps do not apply anymore and instead proposed an elastic model to solve the indeterminacy of contact forces.
However, their approach uses a linearized  formulation~\cite{Haas-Heger2020}, while we propose a non-linear formulation which is harder to optimize but physically valid.
\citet{Abdeetedal2018} extended the approach of  \citet{Bicchi1991} to support kinematic uncertainties.
Both approaches need tactile sensors on all links to be able to estimate the applied external wrench via the measured contact forces.
Instead, our approach only relies on the torques measured in each joint to do so.

\section{Grasp controller}
\label{sec:grasp_controller}

\begin{figure}[htbp]
\centering
\footnotesize
\vspace{0.2cm}
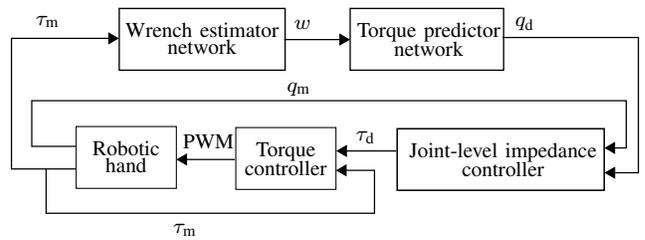%
\caption{Block diagram of our proposed controller. The controller is built on a joint-level impedance controller which commands the underlying torque controller based on the given desired joint angles. The core of our controller consists of the estimation of the external wrench based on the measured joint torques and the subsequent prediction of the desired joint angles which are necessary to counter the external wrench.}
\label{fig:controller}
\vspace{-10pt}
\end{figure}

\begin{figure}[h]
\centering
\begin{minipage}{0.8\linewidth}
\footnotesize
\begingroup%
  \makeatletter%
  \providecommand\color[2][]{%
    \errmessage{(Inkscape) Color is used for the text in Inkscape, but the package 'color.sty' is not loaded}%
    \renewcommand\color[2][]{}%
  }%
  \providecommand\transparent[1]{%
    \errmessage{(Inkscape) Transparency is used (non-zero) for the text in Inkscape, but the package 'transparent.sty' is not loaded}%
    \renewcommand\transparent[1]{}%
  }%
  \providecommand\rotatebox[2]{#2}%
  \newcommand*\fsize{\dimexpr\f@size pt\relax}%
  \newcommand*\lineheight[1]{\fontsize{\fsize}{#1\fsize}\selectfont}%
  \ifx\svgwidth\undefined%
    \setlength{\unitlength}{193.40787272bp}%
    \ifx\svgscale\undefined%
      \relax%
    \else%
      \setlength{\unitlength}{\unitlength * \real{\svgscale}}%
    \fi%
  \else%
    \setlength{\unitlength}{\svgwidth}%
  \fi%
  \global\let\svgwidth\undefined%
  \global\let\svgscale\undefined%
  \makeatother%
  \begin{picture}(1,0.58628258)%
    \lineheight{1}%
    \setlength\tabcolsep{0pt}%
    \put(0,0){\includegraphics[width=\unitlength,page=1]{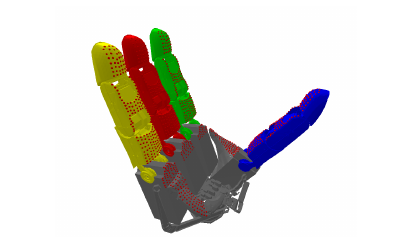}}%
    \put(0.8374836,0.47590996){\makebox(0,0)[lt]{\lineheight{1.25}\smash{\begin{tabular}[t]{l}thumb\end{tabular}}}}%
    \put(0.4375664,0.54130387){\makebox(0,0)[lt]{\lineheight{1.25}\smash{\begin{tabular}[t]{l}forefinger\end{tabular}}}}%
    \put(0.05166477,0.53220109){\makebox(0,0)[lt]{\lineheight{1.25}\smash{\begin{tabular}[t]{l}middlefinger\end{tabular}}}}%
    \put(0.02811576,0.39339){\makebox(0,0)[lt]{\lineheight{1.25}\smash{\begin{tabular}[t]{l}ringfinger\end{tabular}}}}%
    \put(0,0){\includegraphics[width=\unitlength,page=2]{dlr_hand_2.pdf}}%
  \end{picture}%
\endgroup%
\end{minipage}%
\begin{minipage}{0.2\linewidth}
\footnotesize
\begingroup%
  \makeatletter%
  \providecommand\color[2][]{%
    \errmessage{(Inkscape) Color is used for the text in Inkscape, but the package 'color.sty' is not loaded}%
    \renewcommand\color[2][]{}%
  }%
  \providecommand\transparent[1]{%
    \errmessage{(Inkscape) Transparency is used (non-zero) for the text in Inkscape, but the package 'transparent.sty' is not loaded}%
    \renewcommand\transparent[1]{}%
  }%
  \providecommand\rotatebox[2]{#2}%
  \newcommand*\fsize{\dimexpr\f@size pt\relax}%
  \newcommand*\lineheight[1]{\fontsize{\fsize}{#1\fsize}\selectfont}%
  \ifx\svgwidth\undefined%
    \setlength{\unitlength}{48.18897773bp}%
    \ifx\svgscale\undefined%
      \relax%
    \else%
      \setlength{\unitlength}{\unitlength * \real{\svgscale}}%
    \fi%
  \else%
    \setlength{\unitlength}{\svgwidth}%
  \fi%
  \global\let\svgwidth\undefined%
  \global\let\svgscale\undefined%
  \makeatother%
  \begin{picture}(1,2.352947)%
    \lineheight{1}%
    \setlength\tabcolsep{0pt}%
    \put(0,0){\includegraphics[width=\unitlength,page=1]{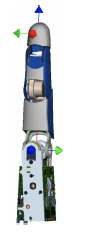}}%
    \put(0.86332459,1.82321724){\color[rgb]{0,0,0}\makebox(0,0)[t]{\lineheight{0.5}\smash{\begin{tabular}[t]{c}$q_4$\end{tabular}}}}%
    \put(-0.45821322,5.97523562){\color[rgb]{0,0,0}\makebox(0,0)[lt]{\begin{minipage}{10.94442974\unitlength}\raggedright \end{minipage}}}%
    \put(-0.10759075,5.39921299){\color[rgb]{0,0,0}\makebox(0,0)[lt]{\begin{minipage}{9.54194018\unitlength}\raggedright \end{minipage}}}%
    \put(0,0){\includegraphics[width=\unitlength,page=2]{finger.pdf}}%
    \put(0.86332459,1.40562817){\color[rgb]{0,0,0}\makebox(0,0)[t]{\lineheight{0.5}\smash{\begin{tabular}[t]{c}$q_3$\end{tabular}}}}%
    \put(0.86332818,0.84470343){\color[rgb]{0,0,0}\makebox(0,0)[t]{\lineheight{0.5}\smash{\begin{tabular}[t]{c}$q_2$\end{tabular}}}}%
    \put(0.07730235,0.84469491){\color[rgb]{0,0,0}\makebox(0,0)[t]{\lineheight{0.5}\smash{\begin{tabular}[t]{c}$q_1$\end{tabular}}}}%
    \put(0,0){\includegraphics[width=\unitlength,page=3]{finger.pdf}}%
  \end{picture}%
\endgroup%
\end{minipage}
\caption{The DLR-Hand II~\cite{Butterfass2001} which we use in our experiments. Each red dot represents a potential contact point. The finger on the right shows the four joints ($q_3$ and $q_4$ are coupled: $q_3 = q_4$)}
\label{fig:hand}
\vspace{-10pt}
\end{figure}

We look at the problem of controlling a robotic hand, that is grasping an object $o$, in a way that the object robustly stays in its initial pose, while changing external wrenches $w \in \mathbb{R}^6$ act upon it.
\cref{fig:hand} shows the torque-controlled \textit{DLR-Hand II}~\cite{Butterfass2001}, that is used.
The hand has $L = 12$ actuated joints, three per finger.
The grasp is defined by a 6D hand pose $H$ and a joint configuration $q_{\text{grasp}} \in \mathbb{R}^{L}$, predicted by our grasping network~\cite{Winkelbauer2022}.
After applying the pre-grasp and going into contact with the object, the grasp controller is activated starting from the resulting joint configuration $q_{\text{init}}$.
Afterward, the grasp is assumed to be quasi-static, which allows us to ignore inertial effects.
The object frame is attached to its center of mass, however this is an arbitrary choice, as our approach is irrespective of the definition of the object frame.

\subsection{Overview}

In the DLR-Hand II, each joint has a torque sensor, allowing the hand to be controlled via a set of torques $\tau \in \mathbb{R}^{L}$.
This is done using a torque controller that adapts the PWM signal of each motor based on the measured torques $\tau_\text{m}$ and a set of desired torques $\tau_\text{d}$.
We base our controller (see \cref{fig:controller}) further on a joint-level impedance controller, which is defined as

\begin{equation}
\tau_{\text{d}} = k_{\text{p}} (q_{\text{d}} - q_{\text{m}}) - k_{\text{d}} \dot q_{\text{m}} \,,
\label{eq:impedance_con}
\end{equation}
with $\dot q_{\text{m}}$ being the measured joint velocities, $q_{\text{d}}$ being the desired, $q_{\text{m}}$ being the measured joint angles and $k_\text{p}$ / $k_\text{d}$ being the stiffness / damping coefficient.
This allows us on the one side to command torques by adapting $q_{\text{d}}$, while still being able to measure changing torques $\tau_{\text{m}}$.
Specifically, in our grasp controller, we first estimate $w$ based on $\tau_{\text{m}}$ and in a second step adjust $q_{\text{d}}$ such that the external wrench is counteracted.
We have these two explicit steps because it is easier to approach each problem isolated and it makes the controller more explainable, as one can inspect the estimated external wrench.

\subsection{Elastic formulation}

As we are targeting all kinds of grasps, including power grasps, we use an elastic formulation to handle the indeterminacy of contact forces~\cite{Bicchi1991, winkelbauer2023}:
The given grasp results in $N$ hand-object contacts, described by their position $p_i$ and object normal $n_i$.
In a stable grasp, the contact forces $\lbrace f_1, ..., f_N \rbrace$ counter the external wrench 
\begin{equation}
  w = -Gf \,,
  \label{eq:wrench_con}
\end{equation}
and cause the torques $\tau$ to be applied at the joints  via

\begin{equation}
\tau = J^\text{T} f \,.
\label{eq:torque_con}
\end{equation}
Here, $G \in \mathbb{R}^{6 \times 3N}$ describes the grasp matrix, $J  \in \mathbb{R}^{L \times 3N}$ the jacobian and $f \in \mathbb{R}^{3N}$ the stacked contact forces.
We model the contact points to be slightly elastic, meaning that a spring is placed at each contact point, whose displacement $d \in \mathbb{R}^{3N}$ is defined by small displacements to the object pose $\Delta x \in \mathbb{R}^6$ and to the joint angles $\Delta q \in \mathbb{R}^{L}$:

\begin{equation}
  d = G^\text{T} \Delta x - J \Delta q + d_{\text{init}}\,.
\label{eq:displacement_con}
\end{equation}
Here, $d_{\text{init}}$ describes the displacement at the starting point of the controller.
Based on the stiffnesses $k_\text{c} \in \mathbb{R}^{N}$ of the contact springs, the contact displacements $d$ relate to the contact forces $f$.
The normal forces $f_\text{n}$ are defined as 

\begin{equation}
  f_{\text{n},i}  = \begin{cases*}
  k_{\text{c}, i} d_{\text{n},i} & if  $n_i \cdot d_{\text{n},i} < 0$  \\
  0 & if $n_i \cdot d_{\text{n},i} \ge 0$
\end{cases*}  \,,
\label{eq:normal_con}
\end{equation}
with $d_{\text{n},i}$ being the the displacement along the contact normal $n_i$.
The tangential forces $f_\text{t}$ are defined as 

\begin{equation}
   f_{\text{t},i} = \begin{cases*}
    k_{\text{c}, i} d_{\text{t},i} & if  $ k_{\text{c}, i}  \norm{d_{\text{t},i}} < \mu \norm{f_{\text{n},i}}$  \\
    \mu \norm{f_{\text{n},i}} \frac{d_{\text{t},i}}{\norm{d_{\text{t},i}}} & else 
\end{cases*} \,,
\label{eq:friction_con}
\end{equation}
with $d_{\text{t},i}$ being the displacement along the contact tangentials and $\mu$ being the friction coefficient.
For further explanations regarding the elastic formulation above, we refer to \citet{winkelbauer2023}.

Due to the usage of the joint-level impedance controller (see \cref{eq:impedance_con}), the relation between torques and joint displacements is further fixed to
 
\begin{equation}
\tau = k_{\text{p}} (q_{\text{d}} - (q_{\text{init}} + \Delta q)) \,.
\label{eq:joint_imp_elastic}
\end{equation}

\subsection{Wrench optimization}
\label{sec:analytical_wrench}

Given measured torques $\tau$ and desired joint angles $q_{\text{d}}$, the external wrench $w$ applied to the object should be determined.
For grasps that contain singularities or where the object is in contact with the static hand base, there can be infinite possible $w$ that fit the given measurements.
To resolve this ambiguity, we desire to find the smallest external wrench $w$.
We found that this improves the stability of the controller as no non-existent wrenches are hallucinated.
In detail, the optimization problem is defined as 

\begin{equation}
\begin{aligned}
\min_{\Delta x, \Delta q} \quad & \ltwo{w_{\text{f}}}\\
\textrm{s.t.} \quad & \textrm{\crefrange{eq:wrench_con}{eq:joint_imp_elastic} are fulfilled,} \\
\end{aligned}
\label{eq:objective_wrench}
\end{equation}
with $w_{\text{f}}$ being the 3D force part of the wrench.

\subsection{Torque optimization}

Based on the estimated external wrench $w$, the desired joint angles $q_{\text{d}}$ are adapted in the second step.
Here, we add the additional constraint

\begin{equation}
\forall l \sum_{i \in \mathcal{F}_l} \mu \norm{f_{\text{n},i}} - \norm{f_{\text{t},i}} > v \,,
\label{eq:margin}
\end{equation}
with $v$ being the minimum friction margin, $l$ being the finger index and $\mathcal{F}_l$ being the set of all contacts placed on the tip of finger $l$.
This makes sure, that each fingertip has contact forces with a minimal margin to the friction cone and therefore the fingers do not slip off the object in the case of a changing external wrench.
Again, usually an infinite amount of possible $q_{\text{d}}$ can be found.
Identical to \citet{winkelbauer2023}, we want to be as efficient as possible and therefore find the $q_{\text{d}}$ which leads to the smallest torques:
So, the optimization problem is defined as 

\begin{equation}
\begin{aligned}
\min_{\Delta x, \Delta q} \quad & \ltwo{\tau} \\
\textrm{s.t.} \quad & \textrm{\crefrange{eq:wrench_con}{eq:joint_imp_elastic} and \cref{eq:margin} are fulfilled} \\
\end{aligned}
\label{eq:objective_torque}
\end{equation}

\section{Learning-based formulation}
\label{sec:learning}

\begin{figure*}[h!]
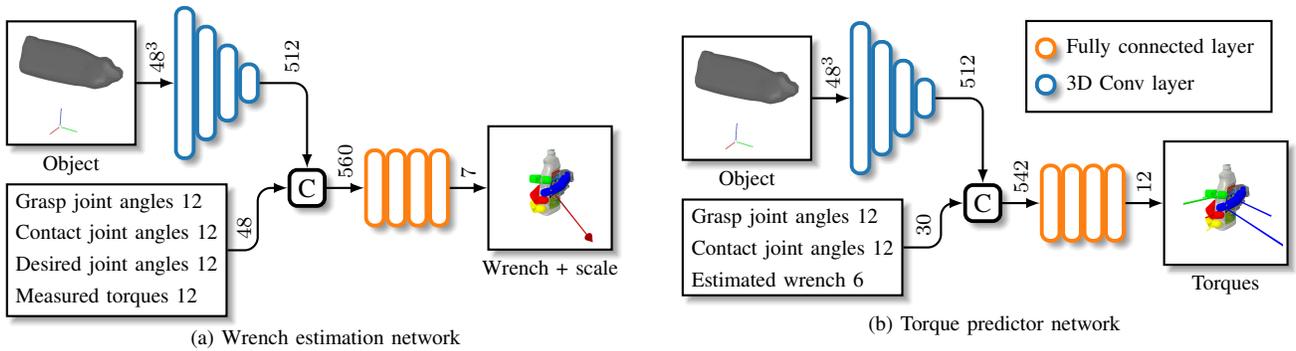

\centering
\subfloat[Wrench estimation network]{%
 \begin{minipage}{0.5\textwidth}
\centering
\vspace{0.01cm}
\input{figures/wrench_network.tex}
\label{fig:network_wrench}
\end{minipage}%
}
\subfloat[Torque predictor network]{%
  \begin{minipage}{0.5\textwidth}
\centering
\vspace{0.01cm}
\input{figures/torque_network.tex}
\label{fig:network_torque}
\end{minipage}%
}
\caption{Architectures of both used networks. The object, represented as a voxelgrid, is encoded using 3D convolutional layers and afterward concatenated with the additional one-dimensional input. A four-layer MLP estimates the external wrench or respectively the desired torques in the end.}
\vspace{-10pt}
\label{fig:networks}
\end{figure*}

The analytical formulation of the controller, described in \cref{sec:grasp_controller}, cannot be run in real-time on the real robot, as both steps require computationally expensive non-linear optimizations (the torque optimization takes ca. \SI{200}{\second} per iteration on a single CPU core).
Further, exact contact points and normals are not known when grasping unknown objects.
That is why we propose to learn these two steps, each using a neural network.
While the offline training of the networks is still computationally expensive, the online inference is fast, which allows us to run the controller in real-time.

\subsection{Wrench estimation}

The wrench estimation network $\mathcal{W}$ learns the mapping 

\begin{equation}
\mathcal{W}: \{o, q_{\text{grasp}}, q_{\text{init}}, q_{\text{d}}, \tau_m\} \rightarrow \{w\} \,.
\end{equation}

\subsubsection{Architecture}
The input to the network consists of the joint angles predicted by the grasping network $q_{\text{grasp}}$, the measured joint angles $q_{\text{init}}$ after applying the pre-grasp, the current desired joint angles $q_{\text{d}}$ , the measured torques $\tau_{\text{m}}$  and the object observation $o$.
The object is transformed into the local coordinate frame of the hand and represented as a $48^3$ voxelgrid, which is available based on the observed depth image through a preceding shape completion step~\cite{humt2023grasping}.
Also, the wrench estimated by the network is represented relative to the coordinate frame of the hand, which makes it easier for the network to generalize across grasps.
As visualized in \cref{fig:network_wrench}, the network consists of four 3D convolutional layers, which encode the object input $o$.
Afterwards, the flattened object embedding is concatenated with the other inputs and given to four fully connected layers which estimate the wrench $w$.

\subsubsection{Training data}
\label{sec:train_wrench}

The results of solving \cref{eq:objective_wrench} depend on the ambiguity in the observed data.
If perfect contact information and torque measurements are given, the wrench can often be estimated accurately.
However, we are interested in the real-world case where neither is available.
That is why we solve the objective \cref{eq:objective_wrench} via the training loss, which considers the actual ambiguity in the data given to the network, and at the same time generate the training data in the following (inverse) way:
First, we sample one external wrench $w$ and one estimated external wrench $w'$.
Using the estimated one, we use the torque prediction network to predict the corresponding desired joint angles $q_{\text{d}} = \mathcal{T}(o, q_{\text{grasp}}, q_{\text{init}}, w')$.
Now, using the equation system \crefrange{eq:wrench_con}{eq:joint_imp_elastic}, we solve for the measured torques $\tau$ when applying the wrench $w$ while using $q_d$.
This results in a dataset of training samples $\{(\hat{q}_{\text{d},1}, \hat{w}_1), \dots, (\hat{q}_{\text{d},T}, \hat{w}_T)\}$ of size $T$.

\subsubsection{Loss}
\label{sec:loss_wrench}
If one would train the network using a standard L2-loss, the network would learn the mean of all wrenches that correspond to a given ambiguous input.
This can lead to severe problems at inference time, as the network could hallucinate wrenches.
As already mentioned in \cref{sec:analytical_wrench}, we therefore always want to estimate the minimal wrench in ambiguous situations.
Therefore, we propose a novel learning procedure:
We extend the output of the network with an explicit estimation of the wrench scale $s \in \mathbb{R}$.
To learn the scale of the smallest wrench corresponding to the given input, we use the loss function

\begin{equation}
\mathcal{L}_{\text{scale}} = \sum_{i=1}^T \mathds{1}_{s_i > \hat{s_i}} \abs{s_i - \hat{s_i}} + \sum_{i=1}^T -t s_i \,,
\end{equation}
with $\hat{s} = \norm{\hat{w}_\text{f}}$ being the ground truth scale.
By setting its derivative to zero, one finds its optimum to be at $\frac{1}{T} \sum_{i=1}^T \mathds{1}_{s_i > \hat{s_i}} = t$.
This means, the optimal $s_i$ is bigger than $t\in \left[ 0, 1 \right]$ part of the ground-truth scales $\hat{s_i}$ corresponding to a given ambiguous input.
The loss function for the wrench estimation is defined as a standard L2-loss with the exception that only training samples are considered where the scale $\hat{s_i}$ is smaller than the predicted one $s_i$:

\begin{equation}
\mathcal{L}_{\text{wrench}} = \sum_{i=1}^T \mathds{1}_{s_i > \hat{s_i}} \ltwo{w_i - \hat{w_i}} \,.
\end{equation}
So, by setting $t$, one can adjust how precise the network should learn the minimal wrench.
A smaller factor $t$ makes the network go closer to the minimal wrench, however, it will also lead to making use of a smaller subset of the training data.
The total loss is defined as $\mathcal{L}_{\text{scale}} + \mathcal{L}_{\text{wrench}}$.

\begin{figure*}[h!]
\centering
\vspace{-0.2cm}
\subfloat[Bowl ($\norm{w_\text{f}}=\SI{5}{\newton}$)]{%
 \includegraphics[width=0.22\linewidth]{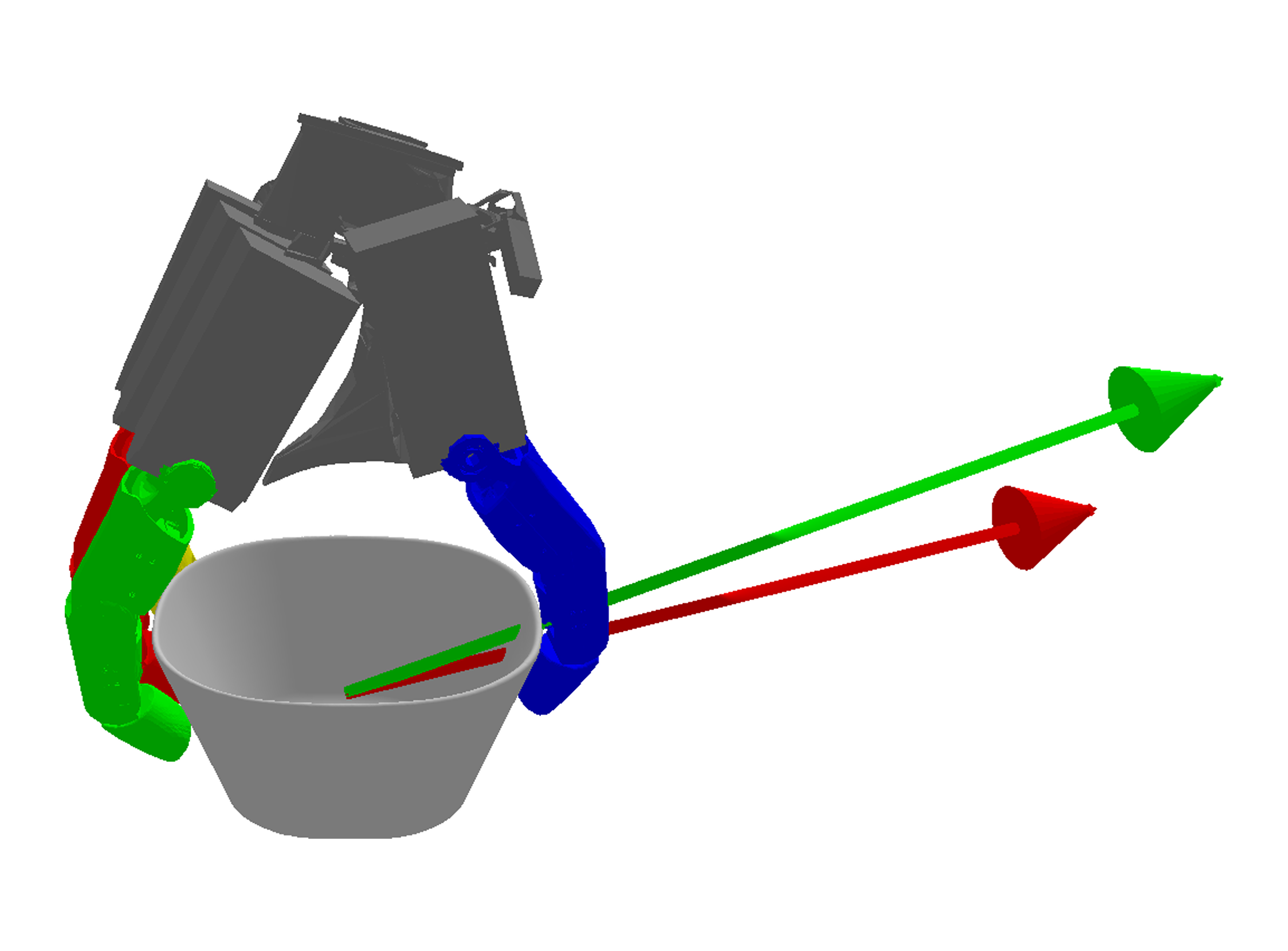}
}
\subfloat[Cereal box ($\norm{w_\text{f}}=\SI{10}{\newton}$)]{%
 \includegraphics[width=0.22\linewidth]{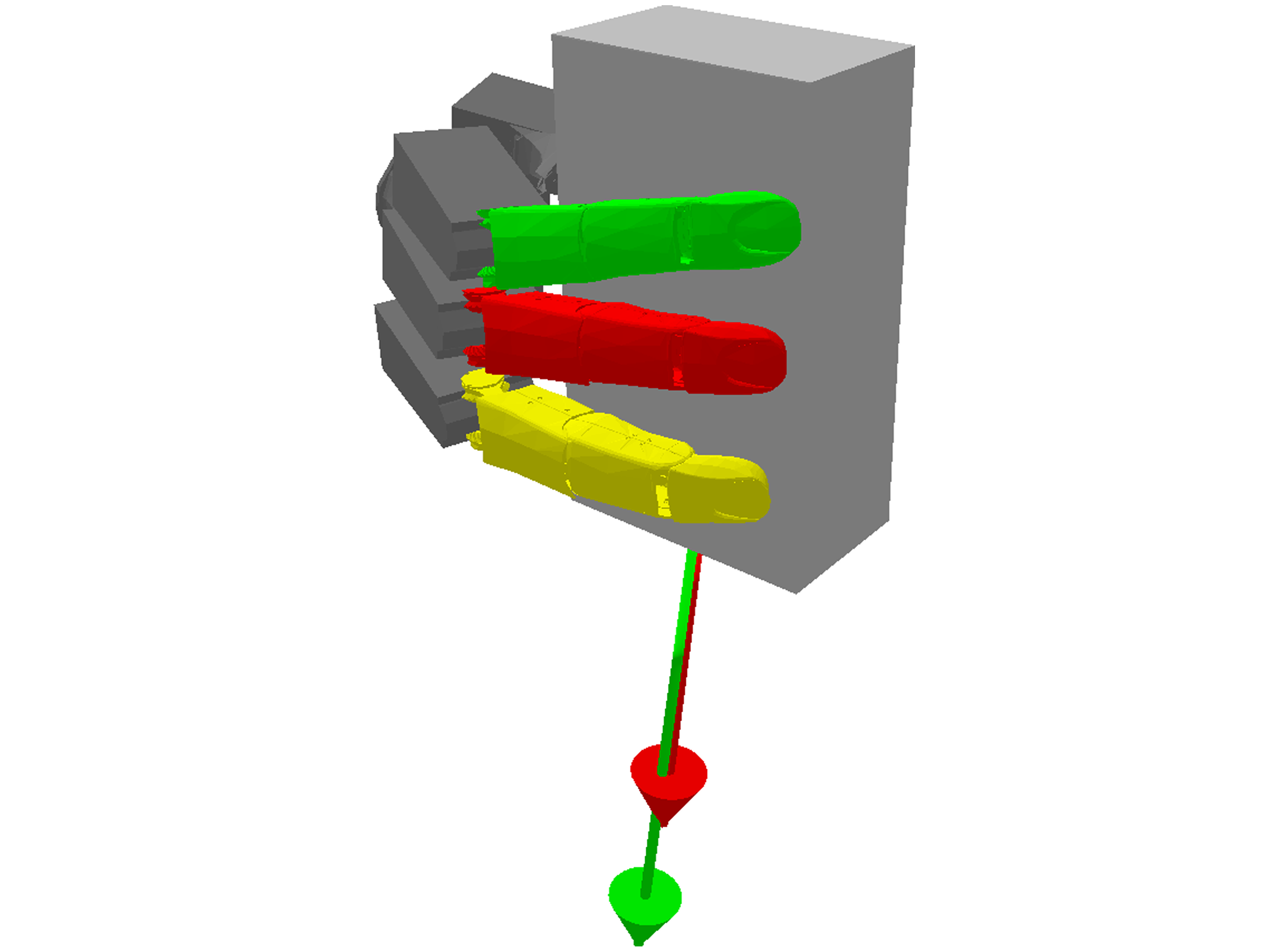}
}
\subfloat[Butter box ($\norm{w_\text{f}}=\SI{5}{\newton}$)]{%
 \includegraphics[width=0.22\linewidth]{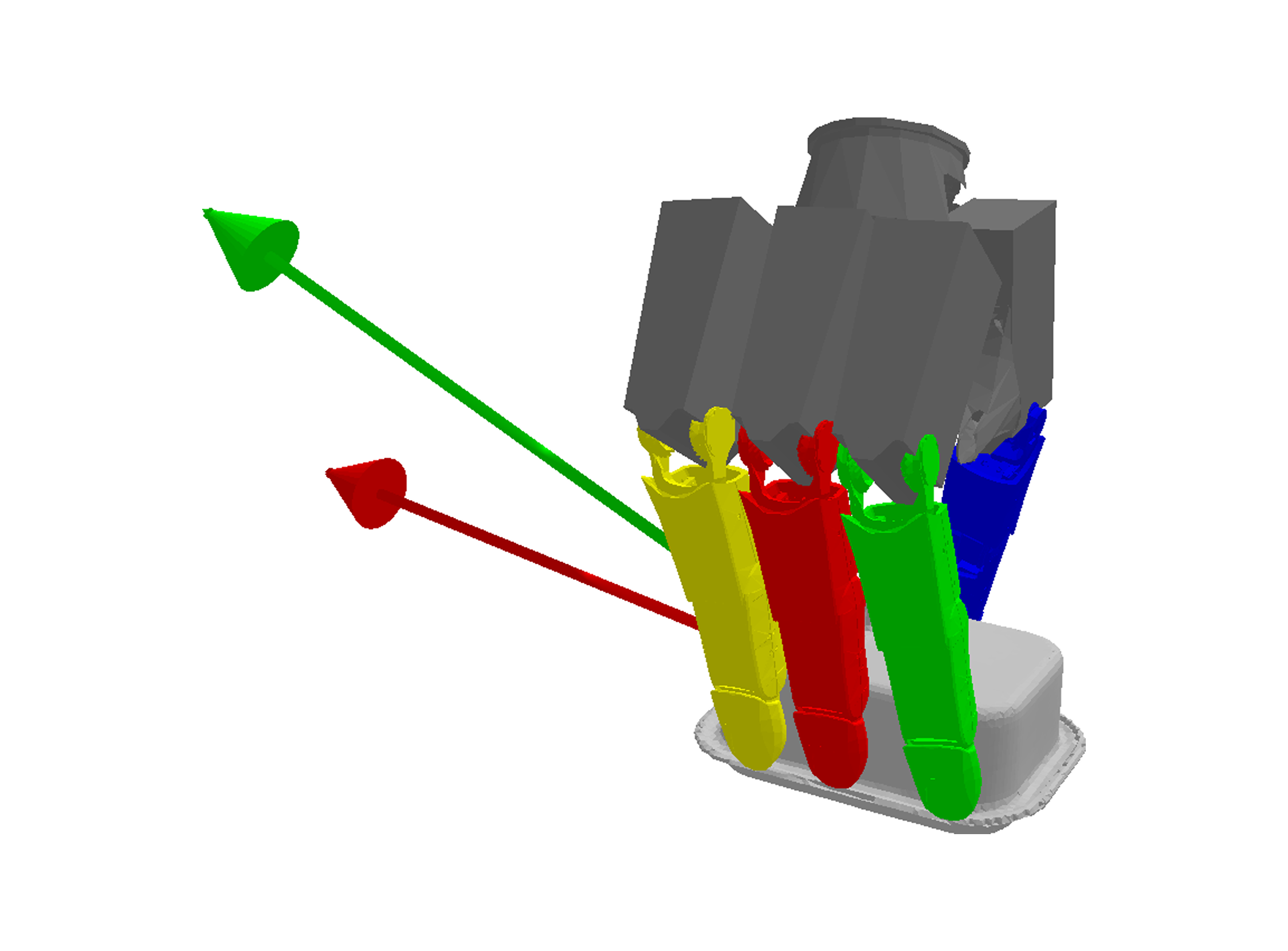}
}
\subfloat[Bleach cleanser ($\norm{w_\text{f}}=\SI{10}{\newton}$)]{%
 \includegraphics[width=0.22\linewidth]{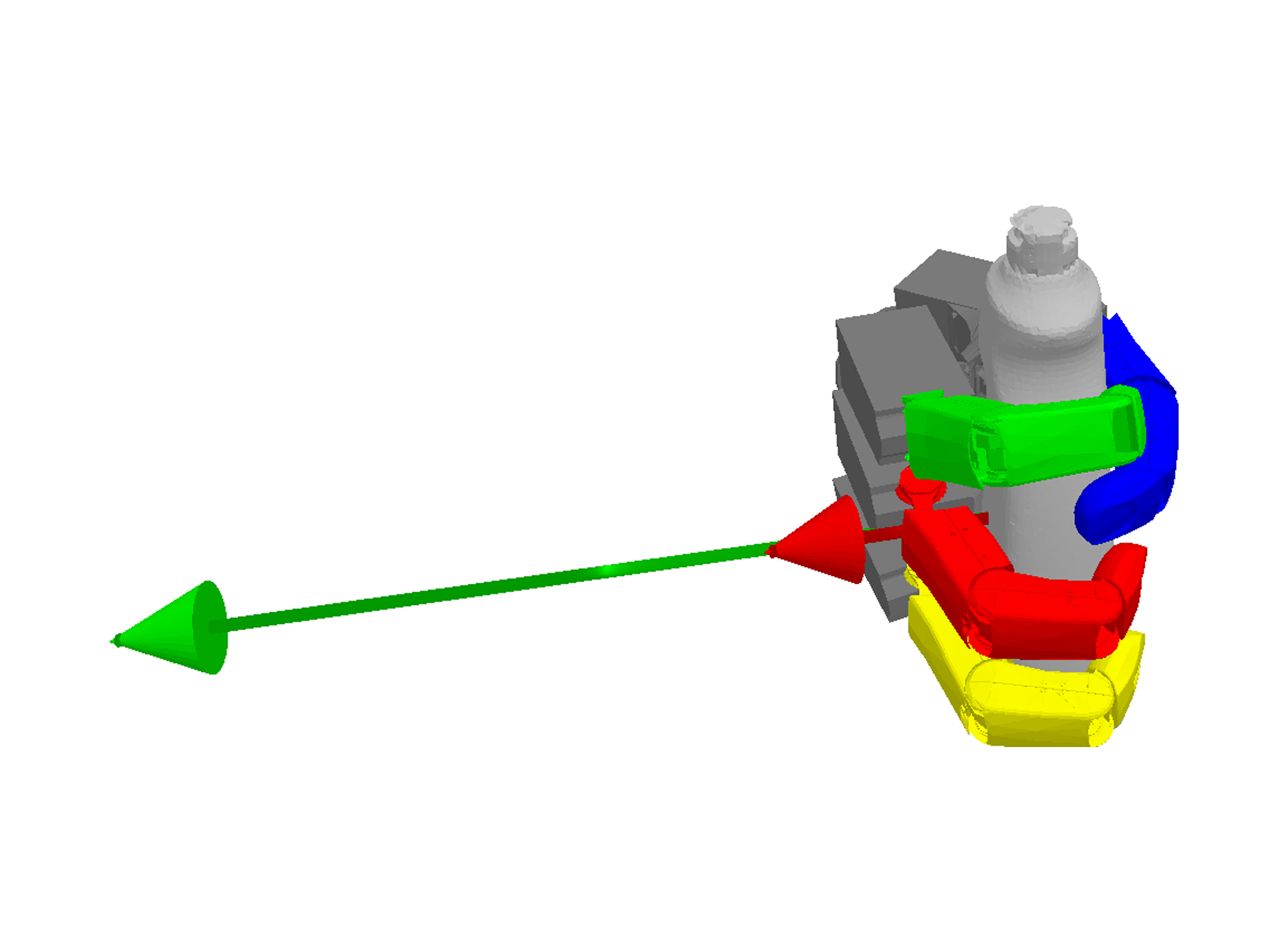}
\label{fig:qualitative_fail}
}
\caption{Qualitative examples of the wrench estimation. Shows the ground truth external wrench in green and the wrench estimated by the network in red. }
\vspace{-10pt}
\label{fig:qualitative}
\end{figure*}

\subsection{Torque prediction}

The torque prediction network $\mathcal{T}$ learns the mapping 

\begin{equation}
\mathcal{T}: \{o, q_{\text{grasp}}, q_{\text{init}}, w\} \rightarrow \{q_{d}\} \,.
\end{equation}

\subsubsection{Architecture}
So, based on the object $o$, the external wrench $w$, the initial joint angles $q_{\text{init}}$ and the grasp joint angles $q_{\text{grasp}}$, the network should predict the smallest desired joint angles $q_d$ which will lead to resisting the external wrench $w$.
Instead of predicting $q_{\text{d}}$ directly, we let the network predict the desired torques $\tau$ and then set the desired joint angles via $q_{\text{d}} = k_{\text{p}} \tau + (q_{\text{init}} + \Delta q)$.
This has the advantage, that the network is independent of  $k_{\text{p}}$.
As $\Delta q$ is in our case relatively insignificant due to $k_{\text{p}} << k_{\text{c}}$, we assume it to be $0$.
However, if necessary one could also let the network additionally predict $\Delta q$ or directly $q_{\text{d}}$.
The architecture of the torque prediction network, visualized in \cref{fig:network_torque}, is congruent to the wrench estimation network.

\subsubsection{Training}
\label{sec:train_torque}
To generate training data, we sample random external wrenches $w$ and then solve the optimization problem \cref{eq:objective_torque}.
Similar to \citet{winkelbauer2023}, we approach the highly non-linear problem using a local gradient-based root solver combined with a global evolutionary optimizer.
This results in a training dataset $\{(\hat{w}_1, \hat{\tau}_1), \dots, (\hat{w}_T, \hat{\tau}_T)\}$.
A standard L2 loss is used to train the network.

\section{Experiments}

\begin{figure}
\centering
\includegraphics[width=0.75\linewidth]{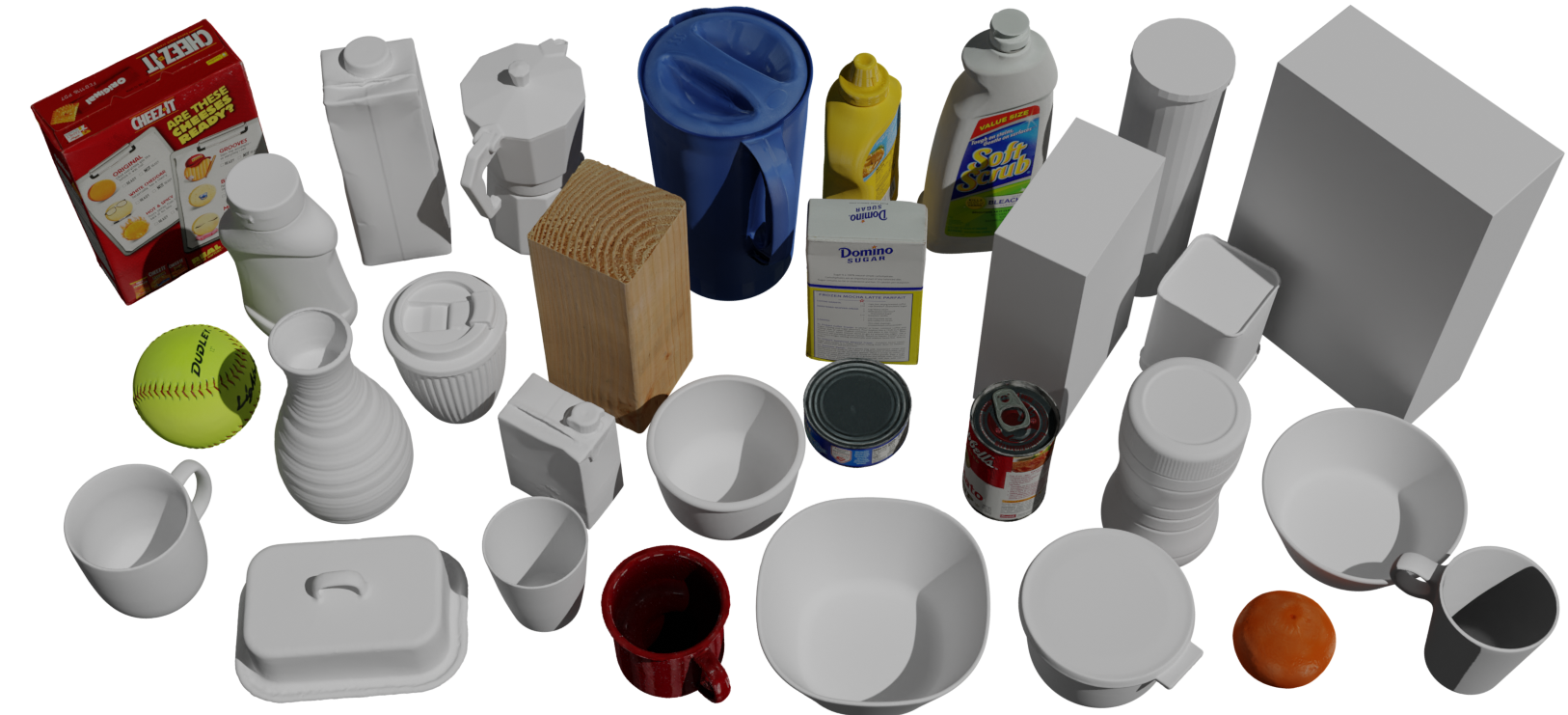}
\caption{The 30 objects used in the evaluation.}
\label{fig:objects}
\vspace{-10pt}
\end{figure}

\subsection{Training}

For generating the training data, we use $\mu = 0.4$,  $k_{\text{c}} = \SI{10000}{\newton\per\meter}$ across all contacts and $k_{\text{p}} = \SI{5}{\newton\meter\per\radian}$.
We generate training data using \num{15000} objects from the ShapeNet dataset~\cite{shapenet2015}.
For each object, we simulate a Kinect-like depth image and apply shape completion to reconstruct the original object~\cite{humt2023grasping}.
Based on that, we use our grasping network~\cite{Winkelbauer2022} to generate \num{1000} grasps for each object, from which we select \num{24}.
This leads to a total dataset of \num{360000} grasps.
For each grasp, we simulate the pre-grasp and afterward gather dense contact points (see potential contact points in \cref{fig:hand}).
For the wrench estimation network, we generate \num{40} training samples per grasp, as described in \cref{sec:train_wrench}.
The resulting dataset of \num{1e7} samples is used to train the network using $t=0.05$ for \num{2.5e6}~iterations.
For the torque prediction network, we generate one training sample per grasp, leading to a dataset size of \num{230000}.
The network is trained for \num{500000}~iterations.
Both networks are trained using the Adam optimizer and a learning rate of \num{3e-4}.

When sampling external wrenches, we use the space of all possible 3D forces and additionally make use of the \textit{object wrench space} (OWS)~\cite{pollard1994}.
In detail, we restrict ourselves to all wrenches that can be caused by a single force being applied to the surface of the object.
This has the advantage, that an intuitive scaling between forces and torques is provided.
We uniformly sample $\norm{w_\text{f}}$ to be between \SI{0}{\newton} and \SI{10}{\newton}.

\subsection{Experimental setup}

To verify our grasp controller, we evaluate it in simulation across 240 grasps on 30 objects (see \cref{fig:objects}).
All grasps and objects are unknown to the networks.
Each grasp is evaluated \num{100} times using different random external wrenches $w$. 
For sampling the wrench $w$ we again take inspiration from the OWS, meaning we sample a point on the surface of the object and apply a force along the object normal.
We use $\norm{w_\text{f}} = \SI{10}{\newton}$, except for precision grasps, where we use $\norm{w_\text{f}} = \SI{5}{\newton}$ to stay inside the torque limits of the joints.
The external wrench is applied to the object by linearly scaling it up from \num{0} to its full size in \SI{4}{\second}.
Afterward, the full wrench is kept constant for \SI{4}{\second}.
We mark a grasp as broken if the object deviates more than \SI{3}{\centi\meter} or \SI{20}{\degree} from its initial pose at any point.
\cref{fig:qualitative} shows four exemplar evaluation runs together with the actual and estimated external wrench.

\begin{figure}
        \includegraphics[width=\linewidth]{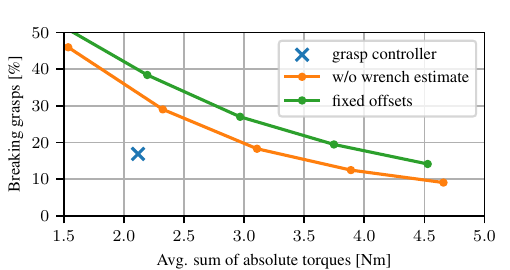}
    \caption{Trade-off between efficiency and breaking grasps. Shows how the scaling factor in both baselines influences the ratio of breaking grasps compared to the required torques and compares it to our proposed controller. Along both axes, lower is better. Shows that our method is more efficient than both tested baselines.}
    \label{fig:breaking}
\vspace{-10pt}
\end{figure}

We compare our grasp controller with two baselines:
One called \textit{fixed offsets}, which uses a fixed set of joint offsets $\Delta q$ across all objects, specifically, \SI{0.2}{\radian} on the $q_2$ and \SI{0.1}{\radian} on the $q_3$ joints.
This represents a common heuristic, where each finger is closed with the same force.
In the other baseline, called \textit{w/o wrench estimate}, we use the torque prediction network to predict for every given grasp a specific $\Delta q = \mathcal{T} (o, q_{\text{init}}, q_{\text{grasp}}, w) - q_{\text{init}}$, using $w = 0$.
During the grasp, $q_\text{d}$ is kept constant.
For both heuristics, we compute $q_\text{d}$ via $q_{\text{d}} = \lambda \Delta q + q_{\text{init}}$ and use the same joint-level impedance controller as in the full grasp controller.
To make a fair comparison, we try out different scaling factors $\lambda \in \lbrace1, 1.5, 2, 2.5, 3\rbrace$ for both baselines.

\subsection{Efficiency}

As shown in \cref{fig:breaking}, when applying our controller on all grasps from the evaluation dataset, \SI{16.9}{\percent} of them break and on average, a torque sum of \SI{2.1}{\newton\meter} is used across all objects.
For comparison when using the \textit{fixed offsets} baseline, the rate of broken grasps is ca. \SI{40}{\percent} when using a scaling factor that results in an equal torque consumption.
If one would like to reach the same success rate, one would need to significantly increase the scale, however, this results in a higher torque usage of ca. \SI{4.1}{\newton\meter}.
The baseline \textit{w/o wrench estimate} performs slightly better, as the joint offsets are specific to each object, however, it leads to the same conclusion:
The baseline is way less efficient than our grasp controller when trying to reach the same success rate.
At a scaling of $1$, the \textit{w/o wrench estimate} baseline equals our grasp controller, when one would keep $q_\text{d}$ constant starting from the first timestep and ignore any changing external wrenches.
In this case, this would lead to a failure rate of \SI{46}{\percent}.
This shows that our controller is actively adapting to changing external wrenches, while still being efficient when no wrench is applied.

\begin{figure}
     \centering 
	\vspace{0.2cm}
     \subfloat[{Translational deviation [cm]}]{
        \centering       
        \includegraphics[width=\linewidth]{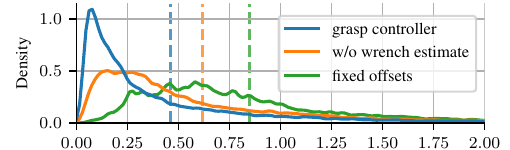}
    	\label{fig:dev_transl}
     }\\
     \subfloat[{Rotational deviation [\degree]}]{    
         \centering
	     \includegraphics[width=\linewidth]{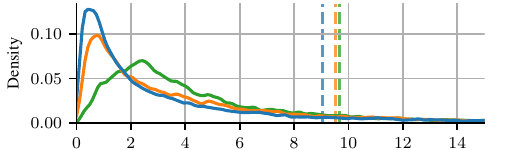}
	    \label{fig:dev_rot}
     }
     
    \caption{Density diagrams showing the maximum object deviation of its initial pose when applying the respective method. Vertical dashed lines show the \SI{90}{\percent} quantile of each method.}
      \label{fig:dev}
\vspace{-10pt}
\end{figure}

\subsection{Object movement}

Another advantage of our controller is that $q_d$ is set in a way that the applied torques counter the external wrench when the object is at its initial pose.
This means, that if the external wrench is estimated correctly, the object barely moves, even under high load.
This is not the case when using one of the baselines.
Here, the joint angles and therefore also the object pose need to deviate until the joint-level impedance controller applies the necessary torque.
This deviation can be reduced by increasing the stiffness $k_p$ of the controller, however, this also leads to more instability.
\cref{fig:dev} shows a density plot over the translational and rotational deviations of the evaluated objects.
Our grasp controller keeps the deviation in \SI{90}{\percent} of the cases below \SI{4.6}{\milli\meter} and \SI{9.0}{\degree} which is in both cases smaller compared to the two baselines.

\subsection{Minimal wrench estimation}

\begin{figure}[h!]
\centering
\subfloat[Grasp]{%
 \includegraphics[width=0.32\linewidth]{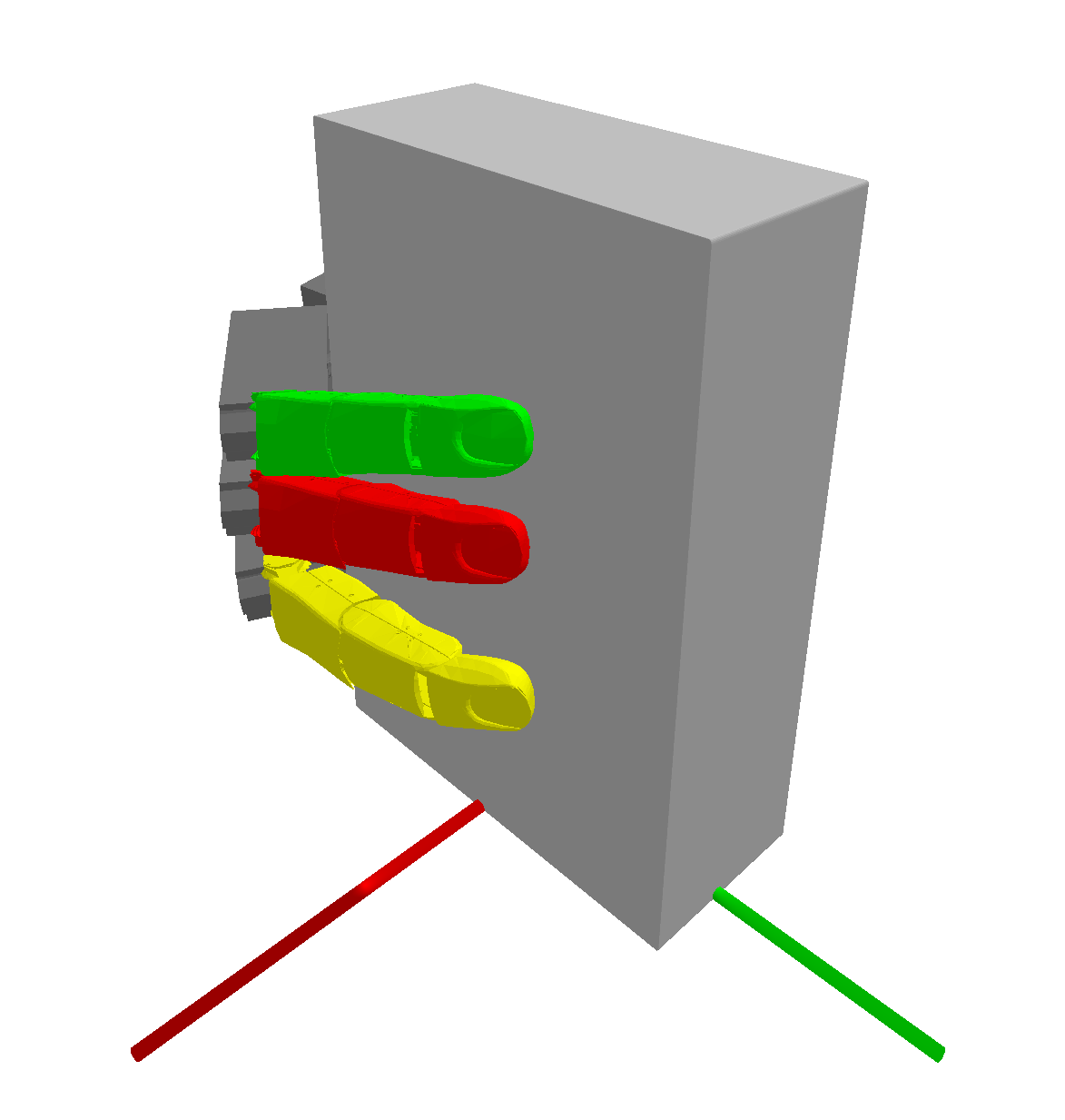}
}
\subfloat[Heatmap]{%
 \includegraphics[width=0.32\linewidth]{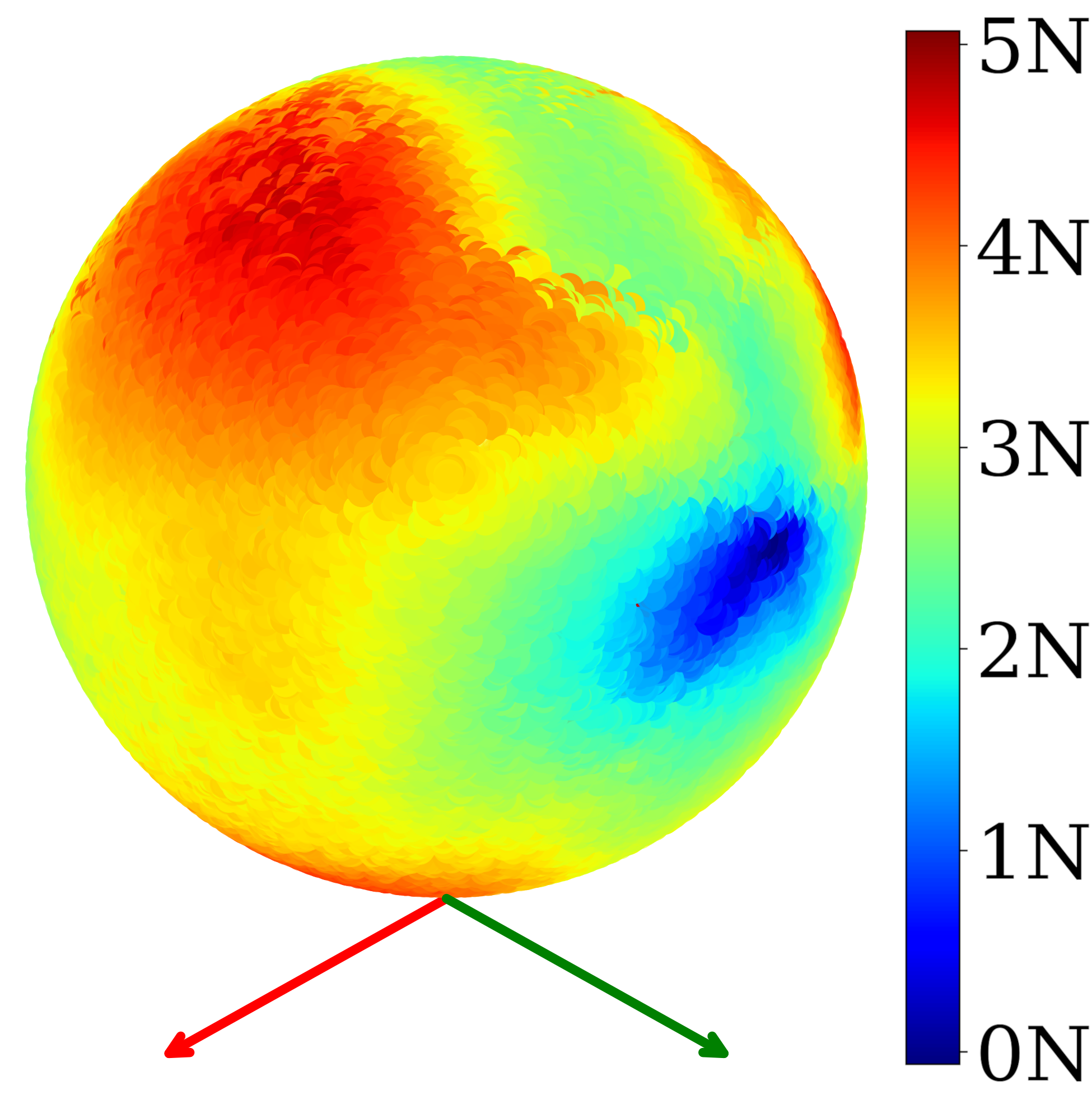}
\label{fig:min_wrench_heat}
}
\subfloat[Standard L2-loss]{%
 \includegraphics[width=0.32\linewidth]{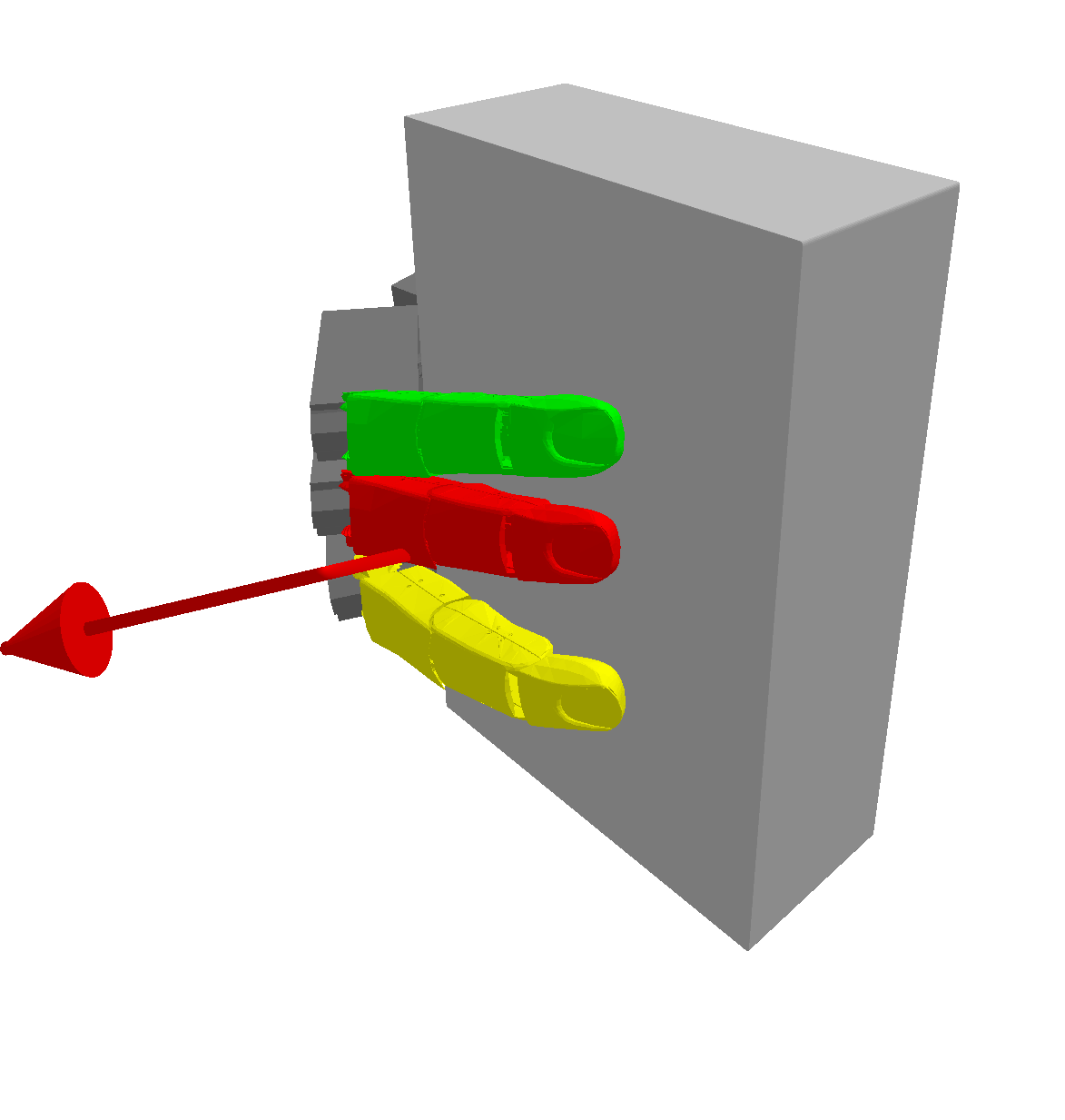}
\label{fig:min_wrench_hallu}
}
\caption{Visualizes the effect of the minimal-wrench loss. For 3D external forces with $\norm{w_\text{f}} = \SI{5}{\newton}$ applied to the object, we let the network estimate them based on the measured torques. The predicted wrench norm in Newton is visualized as a heatmap on a sphere. In \cref{fig:min_wrench_hallu} the hallucinated wrench predicted by a network trained with a standard L2 loss is shown.}
\vspace{-10pt}
\label{fig:min_wrench}
\end{figure}

\begin{figure*}[ht!]
     \centering    
	\vspace{0.2cm}
     \begin{minipage}{4in}
     \subfloat[Butter box]{
        \includegraphics[width=\linewidth]{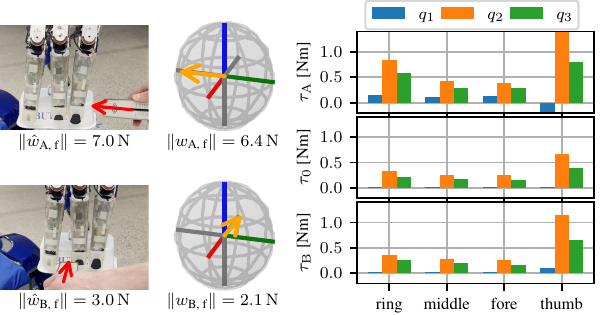}
    	\label{fig:real_butter}
     }
     \end{minipage}%
     \begin{minipage}{1.48in}     
     \subfloat[Cracker box]{
       \includegraphics[width=\linewidth]{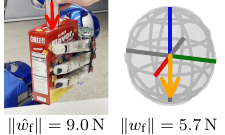}
    	\label{fig:real_cracker}
     }\\%
     \subfloat[Small cup]{
       \includegraphics[width=\linewidth]{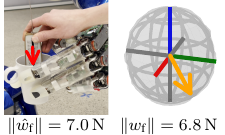}
    	\label{fig:real_cup}
     }    
     \end{minipage}%
     \begin{minipage}{1.48in}     
     \subfloat[Bleach]{
       \includegraphics[width=\linewidth]{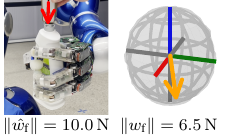}
    	\label{fig:real_bleach}
     }\\%
     \subfloat[Bowl]{
       \includegraphics[width=\linewidth]{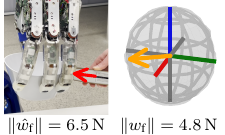}
    	\label{fig:real_bowl}
     }       
     \end{minipage}
    \caption{Our grasp controller applied on the real robot across various objects. The applied forces are highlighted via red arrows and the normalized estimated wrenches are visualized as orange arrows next to each grasp. \cref{fig:real_butter} shows additionally the predicted torques: $\tau_0$ when no wrench is applied , $\tau_\text {A}$ when the first shown wrench is applied and $\tau_\text {B}$ when the second shown wrench is applied.}
      \label{fig:real}
\vspace{-10pt}
\end{figure*}

In the first step of our controller, we estimate the external wrench being applied to the object.
As explained, we decided to always learn the minimal wrench explaining the given measurements. 
Figure \cref{fig:min_wrench} shows what the network trained using the minimal wrench loss predicts for a given grasp on a cereal box.
Here, we sample random 3D forces with norm \SI{5}{\newton}, we use $q_{\text{d}} = \mathcal{T}(o, q_{\text{init}}, q_{\text{grasp}}, 0)$, we solve for the resulting measured torques $\tau$ using \crefrange{eq:wrench_con}{eq:joint_imp_elastic} and then let the network estimate the wrench $w$ again based on $\tau$.
In \cref{fig:min_wrench_heat} we visualize the norm of the predicted forces $\norm{w_\text{f}}$ using a heatmap projected on a sphere.
It is obvious that the predicted force gets toward zero when approaching forces parallel to the fingers along the y-axis.
Due to the singularity, these wrenches do not result in any significant torques measured at the joints. 
Therefore, the network has learned to not predict any wrench at all along this axis.
We further evaluate a network trained with a usual L2-loss:
\cref{fig:min_wrench_hallu} shows the wrench that is estimated by such a network when no external wrench is actually applied.
The network is hallucinating here an external force of size \SI{2.8}{\newton}.
In the same situation, our network trained with the minimal-wrench loss predicts an external force of norm \SI{0.2}{\newton}.

\subsection{Evaluation on the real robot}
\label{sec:exp_real}

We evaluate the proposed grasp controller on our humanoid \textit{Agile Justin}~\cite{Baeuml2014} using our grasping pipeline:
First, the object is observed in the form of a single depth image using the robot's Kinect camera.
Based on the observed incomplete point cloud, we apply our shape completion and grasp prediction networks to find a stable grasp suitable for the given object~\cite{humt2023grasping}.
After approaching the grasp pose using our learning-based motion planner~\cite{Tenhumberg2022}, the fingers are closed until they get into contact with the object and then the hand lifts up the object.
Now, the grasp controller is activated.
\cref{fig:real} shows the estimated external wrench across various grasps.
It can be seen that the estimated wrench fits the actual one in all cases.
Also, \cref{fig:real_butter} shows the predicted torques:
For wrench $\text{A}$, the torques at $q_1$ counter the external wrench, while the ones at $q_2$ and $q_3$ increase the normal forces such that the contacts do not break.
For wrench $\text{B}$, the torques applied at the three fingers decrease and the ones applied at the thumb increase, while the torques at $q_1$ stay constant.
This shows that the predicted torques actively counter the estimated external wrench.
In the experiments, the controller ran with a frequency of \SI{40}{\hertz}.
However, also higher frequencies are possible, as one controller iteration takes \SI{6}{\milli\second} on a \textit{NVIDIA V100}.

\subsection{Discussion}

Due to the physical limitations of the sensors used in the proposed controller, our approach has the following properties:
Singularities in the finger kinematics can lead to the wrench not being measured at the torque sensors of the joint angles and therefore the wrench not being estimated correctly.
Another reason for not estimating the wrench correctly are contact points with the fixed hand base, where contact forces are unknown.
An example is shown in \cref{fig:qualitative_fail}.
Furthermore, ambiguity also arises from the fact that we do not know the exact contact positions/normals and the networks have to estimate them based on the observed and reconstructed object shape together with the joint positions.

\section{Conclusions}

In this work, we proposed a novel torque-based grasp controller that is able to robustly hold unknown objects.
The controller only uses the measured joint torques and therefore does not require extensive object movement, which allows us to support power grasps with multi-contacts.
We provided a full analytical formulation and proposed a learning-based adaptation, which decreases the cycle time to \SI{6}{\milli\second} and thus allows real-time performance.
In a thorough evaluation in a physics-based simulator,  
we showed that the controller is more efficient and induces less involuntary object movement compared to the two tested baselines.
We further showed the importance of the proposed minimal-wrench loss which makes sure the neural network does not make the controller unstable by hallucinating non-existent external wrenches.
The controller has been shown to generalize to the real robot, where we showed that it is able to estimate applied wrenches and adapt the joint torques accordingly.
Specifically, the controller was tested on our DLR-Hand II across various objects and grasp types.
We leave it up for future work to reduce these physical limitations of our approach by incorporating more sensor modalities.

\footnotesize
\bibliographystyle{IEEEtranN-modified}
\bibliography{IEEEabrv,bibliography}

\begin{thebibliography}{24}
\providecommand{\natexlab}[1]{#1}
\providecommand{\url}[1]{#1}
\csname url@samestyle\endcsname
\providecommand{\newblock}{\relax}
\providecommand{\bibinfo}[2]{#2}
\providecommand{\BIBentrySTDinterwordspacing}{\spaceskip=0pt\relax}
\providecommand{\BIBentryALTinterwordstretchfactor}{4}
\providecommand{\BIBentryALTinterwordspacing}{\spaceskip=\fontdimen2\font plus
\BIBentryALTinterwordstretchfactor\fontdimen3\font minus
  \fontdimen4\font\relax}
\providecommand{\BIBforeignlanguage}[2]{{%
\expandafter\ifx\csname l@#1\endcsname\relax
\typeout{** WARNING: IEEEtranN.bst: No hyphenation pattern has been}%
\typeout{** loaded for the language `#1'. Using the pattern for}%
\typeout{** the default language instead.}%
\else
\language=\csname l@#1\endcsname
\fi
#2}}
\providecommand{\BIBdecl}{\relax}
\BIBdecl

\bibitem[Lu et~al.(2020)Lu, van~der Merwe, and Hermans]{Lu2020}
Q.~Lu, M.~van~der Merwe, and T.~Hermans, ``{Multi-Fingered Active Grasp
  Learning},'' in \emph{IEEE International Conference on Intelligent Robots and
  Systems}, 2020.

\bibitem[Wu et~al.(2020)Wu, Akinola, Gupta, Xu, Varley, Watkins-Valls, and
  Allen]{Wu2020}
B.~Wu \emph{et~al.}, ``{Generative Attention Learning: a “GenerAL”
  framework for high-performance multi-fingered grasping in clutter},''
  \emph{Autonomous Robots}, vol.~44, no.~6, 2020.

\bibitem[Aktas et~al.(2019)Aktas, Zhao, Kopicki, Leonardis, and
  Wyatt]{Aktas2019}
U.~R. Aktas \emph{et~al.}, ``{Deep dexterous grasping of novel objects from a
  single view},'' \emph{IEEE Transactions on Robotics}, 2019.

\bibitem[Mandikal and Grauman(2021)]{Mandikal2021}
P.~Mandikal and K.~Grauman, ``{DexVIP : Learning Dexterous Grasping with Human
  Hand Pose Priors from Video},'' in \emph{International Conference on Robot
  Learning}, 2021.

\bibitem[Winkelbauer et~al.(2022)Winkelbauer, Bäuml, Humt, Thuerey, and
  Triebel]{Winkelbauer2022}
D.~Winkelbauer \emph{et~al.}, ``A two-stage learning architecture that
  generates high-quality grasps for a multi-fingered hand,'' in \emph{IEEE
  International Conference on Intelligent Robots and Systems}, 2022.

\bibitem[Schneider and Cannon(1992)]{Schneider1992}
S.~A. Schneider and R.~H. Cannon, ``{Object Impedance Control for Cooperative
  Manipulation: Theory and Experimental Results},'' \emph{IEEE Transactions on
  Robotics and Automation}, vol.~8, no.~3, 1992.

\bibitem[Wimboeck et~al.(2006)Wimboeck, Ott, and Hirzinger]{Wimboeck2006}
T.~Wimboeck, C.~Ott, and G.~Hirzinger, ``{Passivity-based object-level
  impedance control for a multifingered hand},'' in \emph{IEEE International
  Conference on Intelligent Robots and Systems}, 2006.

\bibitem[Wimb{\"{o}}ck et~al.(2012)Wimb{\"{o}}ck, Ott, Albu-Sch{\"{a}}ffer, and
  Hirzinger]{Wimbock2012}
T.~Wimb{\"{o}}ck, C.~Ott, A.~Albu-Sch{\"{a}}ffer, and G.~Hirzinger,
  ``{Comparison of object-level grasp controllers for dynamic dexterous
  manipulation},'' \emph{International Journal of Robotics Research}, vol.~31,
  no.~1, 2012.

\bibitem[Pfanne et~al.(2020)Pfanne, Chalon, Stulp, Ritter, and
  Albu-Sch{\"{a}}ffer]{Pfanne2020}
M.~Pfanne \emph{et~al.}, ``{Object-Level Impedance Control for Dexterous
  In-Hand Manipulation},'' \emph{IEEE Robotics and Automation Letters}, vol.~5,
  no.~2, 2020.

\bibitem[Williams and Khatib(1993)]{Williams1993}
D.~Williams and O.~Khatib, ``{Virtual linkage: A model for internal forces in
  multi-grasp manipulation},'' in \emph{IEEE International Conference on
  Robotics and Automation}, 1993.

\bibitem[Su et~al.(2015)Su, Hausman, Chebotar, Molchanov, Loeb, Sukhatme, and
  Schaal]{Su2015}
Z.~Su \emph{et~al.}, ``{Force estimation and slip detection/classification for
  grip control using a biomimetic tactile sensor},'' in \emph{IEEE
  International Conference on Humanoid Robots}, 2015.

\bibitem[Wettels et~al.(2009)Wettels, Parnandi, Moon, Loeb, and
  Sukhatme]{Wettels2009}
N.~Wettels \emph{et~al.}, ``{Grip control using biomimetic tactile sensing
  systems},'' \emph{IEEE Transactions on Mechatronics}, vol.~14, no.~6, 2009.

\bibitem[Romano et~al.(2011)Romano, Hsiao, Niemeyer, Chitta, and
  Kuchenbecker]{Romano2011}
J.~M. Romano \emph{et~al.}, ``{Human-inspired robotic grasp control with
  tactile sensing},'' vol.~27, no.~6, 2011.

\bibitem[Deng et~al.(2020)Deng, Jonetzko, Zhang, and Zhang]{Deng2020}
Z.~Deng, Y.~Jonetzko, L.~Zhang, and J.~Zhang, ``{Grasping force control of
  multi-fingered robotic hands through tactile sensing for object
  stabilization},'' \emph{Sensors (Switzerland)}, vol.~20, no.~4, 2020.

\bibitem[Bicchi(1991)]{Bicchi1991}
A.~Bicchi, ``{Analysis and control of power grasping},'' in \emph{IEEE
  International Conference on Intelligent Robots and Systems}, 1991.

\bibitem[Haas-Heger and Ciocarlie(2020)]{Haas-Heger2020}
M.~Haas-Heger and M.~Ciocarlie, ``{Accurate Energetic Constraints for Passive
  Grasp Stability Analysis},'' \emph{IEEE Transactions on Robotics}, vol.~36,
  2020.

\bibitem[Abdeetedal et~al.(2018)Abdeetedal, Rezaee, Talebi, and
  Abdollahi]{Abdeetedal2018}
M.~Abdeetedal, H.~Rezaee, H.~A. Talebi, and F.~Abdollahi, ``{Optimal adaptive
  Jacobian internal forces controller for multiple whole-limb manipulators in
  the presence of kinematic uncertainties},'' \emph{Mechatronics}, vol.~53,
  2018.

\bibitem[Butterfaß et~al.(2001)Butterfaß, Grebenstein, Liu, and
  Hirzinger]{Butterfass2001}
J.~Butterfaß, M.~Grebenstein, H.~Liu, and G.~Hirzinger, ``{DLR-H}and {II}:
  Next generation of a dextrous robot hand,'' in \emph{IEEE International
  Conference on Robotics and Automation}, 2001.

\bibitem[Winkelbauer et~al.(2023)Winkelbauer, B{\"a}uml, and
  Triebel]{winkelbauer2023}
D.~Winkelbauer, B.~B{\"a}uml, and R.~Triebel, ``Learning-based real-time torque
  prediction for grasping unknown objects with a multi-fingered hand,'' in
  \emph{IEEE International Conference on Intelligent Robots and Systems}, 2023.

\bibitem[Humt et~al.(2023)Humt, Winkelbauer, Hillenbrand, and
  B{\"a}uml]{humt2023grasping}
M.~Humt, D.~Winkelbauer, U.~Hillenbrand, and B.~B{\"a}uml, ``Combining shape
  completion and grasp prediction for fast and versatile grasping with a
  multi-fingered hand,'' in \emph{IEEE International Conference on Humanoid
  Robots}, 2023.

\bibitem[Chang et~al.(2015)Chang, Funkhouser, Guibas, Hanrahan, Huang, Li,
  Savarese, Savva, Song, Su, Xiao, Yi, and Yu]{shapenet2015}
A.~X. Chang \emph{et~al.}, ``{ShapeNet: An Information-Rich 3D Model
  Repository},'' no. arXiv:1512.03012 [cs.GR], 2015.

\bibitem[Pollard(1994)]{pollard1994}
N.~S. Pollard, ``Parallel methods for synthesizing whole-hand grasps from
  generalized prototypes,'' Ph.D. dissertation, 1994.

\bibitem[Bäuml et~al.(2014)Bäuml, Hammer, Wagner, Birbach, Gumpert, Zhi,
  Hillenbrand, Beer, Friedl, and Butterfass]{Baeuml2014}
B.~Bäuml \emph{et~al.}, ``Agile justin: An upgraded member of dlr's family of
  lightweight and torque controlled humanoids,'' in \emph{IEEE International
  Conference on Robotics and Automation}, 2014.

\bibitem[Tenhumberg et~al.(2022)Tenhumberg, Burschka, and
  Bäuml]{Tenhumberg2022}
J.~Tenhumberg, D.~Burschka, and B.~Bäuml, ``Speeding up optimization-based
  motion planning through deep learning,'' in \emph{IEEE International
  Conference on Intelligent Robots and Systems}, 2022.

\end{thebibliography}

\end{document}